\documentclass[11pt,a4paper]{article}
\usepackage[hyperref]{acl2020}
\usepackage{graphicx} 
\usepackage{times}
\usepackage{amsmath}

\usepackage{latexsym}

\usepackage{microtype}

\aclfinalcopy 

\title{Talk to Papers: Bringing Neural Question Answering to Academic Search}
\author{Tianchang Zhao and Kyusong Lee \thanks{Both authors contributed equally} \\
  SOCO AI \\
  \texttt{\{tianchez,kyusongl\}@soco.ai} }

\date{}

\begin{document}
\maketitle
\begin{abstract}
We introduce Talk to Papers\footnote{\url{https://ask.soco.ai}}, which exploits the recent open-domain question answering (QA) techniques to improve the current experience of academic search. It's designed to enable researchers to use natural language queries to find precise answers and extract insights from a massive amount of academic papers. We present a large improvement over classic search engine baseline on several standard QA datasets, and provide the community a collaborative data collection tool to curate the first natural language processing research QA dataset via a community effort.
\end{abstract}

\section{Introduction}
Natural language processing (NLP) is one of the fastest growing field in computational linguistics and artificial intelligence, e.g. ACL has experienced a 140\% growth from 2017 (1419 submissions) to 2020 (3429 submissions). Plus, there are more than 4000 pre-prints published at ArXiv in 2019. As a result, it has become increasingly stressful for researchers to keep up with the evolution of new methods. Today, the common way for researchers to find relevant papers is via searching keywords in Google Scholar\footnote{\url{https://scholar.google.com/}} or Semantic Scholar\footnote{\url{https://www.semanticscholar.org/}}. Although these search engines are great at curating all the papers, they are limited in the following ways: (1) they are based on classic information retrieval methods, and do not handle natural language queries well, e.g. what effects can we get from label smoothing? (2) they are designed to find relevant documents (title and abstract) instead of direct answers to users' questions. Often researchers are looking for answers on specific research questions, e.g, \textit{how to prevent posterior collapse in VAE?} or \textit{how much is it to label sentences via crowdsourcing?} With current search engine, it requires one to read several papers to find these answer. Therefore, it is necessary to create better tools for researchers to find answers from the scientific publications in a more efficient manner.

Meanwhile, machine reading comprehension (MRC), aka question answering (QA) has advanced significantly. Pretrained and then fine-tuned transformer models~\cite{devlin2018bert} have surpassed human performance on a number of datasets, e.g. SQuAD~\cite{rajpurkar2016squad}. Further, \citet{chen2017reading} extended single document MRC to machine reading at scale (MRS), combining the challenges of document retrieval with reading comprehension. Their open-domain QA system is able to find precise answers from millions of unstructured documents using natural language queries and has successfully been applied to the entire Wikipedia which contains more than 5 million articles.

The goal of Talk to Papers is to create a new way of finding answers from scientific publications and advance QA research. Concretely, we \textbf{first} adapted MRS techniques to create a conversational search portal that enable users to ask natural language questions to find precise answers and extract insights from the last 3 year papers published in top-tier NLP conferences, including ACL, NAACL, EMNLP and etc. \textbf{Second}, an initial corpus on these papers is collected and will be released as a publicly available dataset for QA research. We also developed a collaborative annotation toolkit that enable any researcher to contribute to this dataset so that more potential answers from these papers can be annotated. The annotation results will be fed back to the QA corpus after manual validation. 


\section{Related Work}
\begin{figure*}[t]
\centering
\includegraphics[width=16cm]{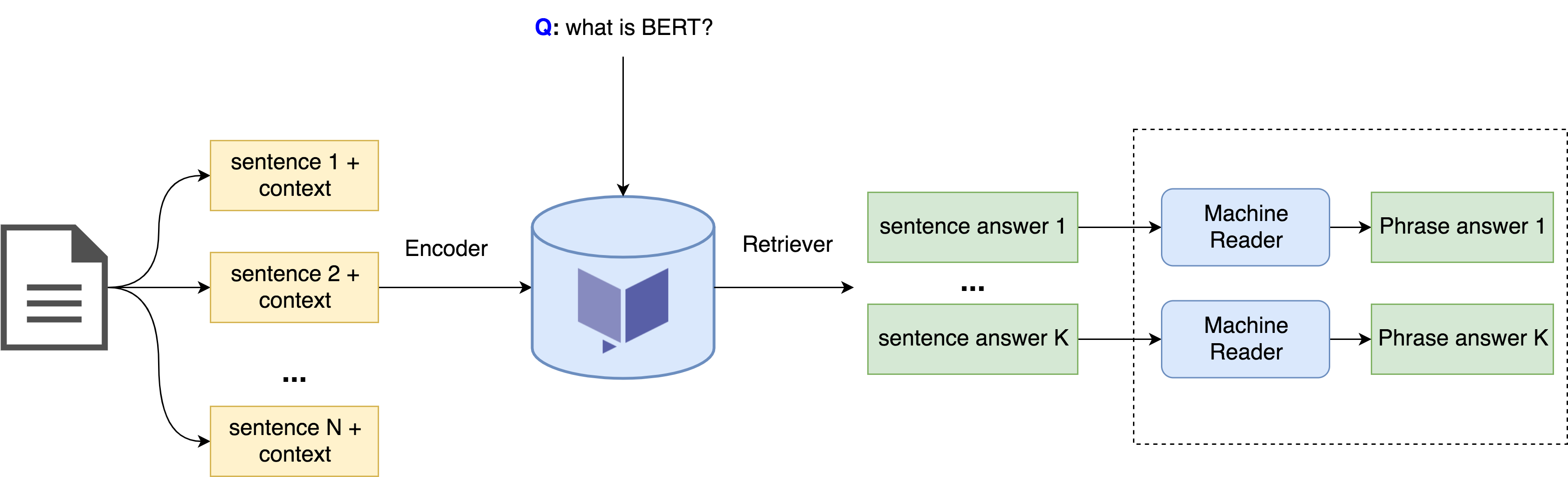}
\caption{Overall workflow of the proposed SOCO framework. The machine reading step in the dashed box is optional.}
\label{fig:overall}
\end{figure*}

Developing question answering system with text knowledge base has been studied for decades~\cite{voorhees1999trec}. Many of the classic system as well as recent MRC-based open-domain QA systems have relied a pipeline approach~\cite{ferrucci2010building,chen2017reading,lee2018ranking,yang2019end}: (1) a information retrieval-based retriever module first finds relevant passages from all the documents and then (2) a reader component (computationally more expensive) extracts precise answer spans from the retrieved passages. \citet{yang2019end} has shown that using paragraphs as the unit of passage outperform sentences or documents. \citet{lee2019latent} proposes a trainable first-stage retriever that improves the recall performance.

Pipeline-based system often suffer from error propagation~\cite{zhao2016towards}. Thus another line of research has been finding an end-to-end approach that enable precise-answer extraction from the entire dataset instead of only the output from the first-stage retriever. \citet{seo2019real} introduced the phrase level representation model that index every potential answer span as vector representation and exploited approximate nearest neighbour (ANN) methods to retrieve the final answer span directly from a large vector index~\cite{slaney2008locality}. \citet{ahmad2019reqa} argued that phrase-level answer may not always be required or preferred. Instead they proposed to find the right ``sentence'' as an answer from large body of text, and used universal sentence encoder~\cite{cer2018universal} to retrieved the correct sentence given a question. 

Our approach follows the sentence-level QA system from~\cite{ahmad2019reqa} for two reasons: (1) answers to many research questions cannot be covered in a short phrase-level span, and a sentence answer can provide more context to deliver relevant solutions. (2) our preliminary study found that it is important to have a trainable retriever that goes beyond TF-IDF keyword matching to ensure enough recall in the paper domain. Nonetheless, we keep a machine reader as optional post-process to extract phrase-level span from the sentences. 

\section{The Proposed QA Toolkits: SOCO}
\begin{figure*}[t]
\centering
\includegraphics[width=16cm]{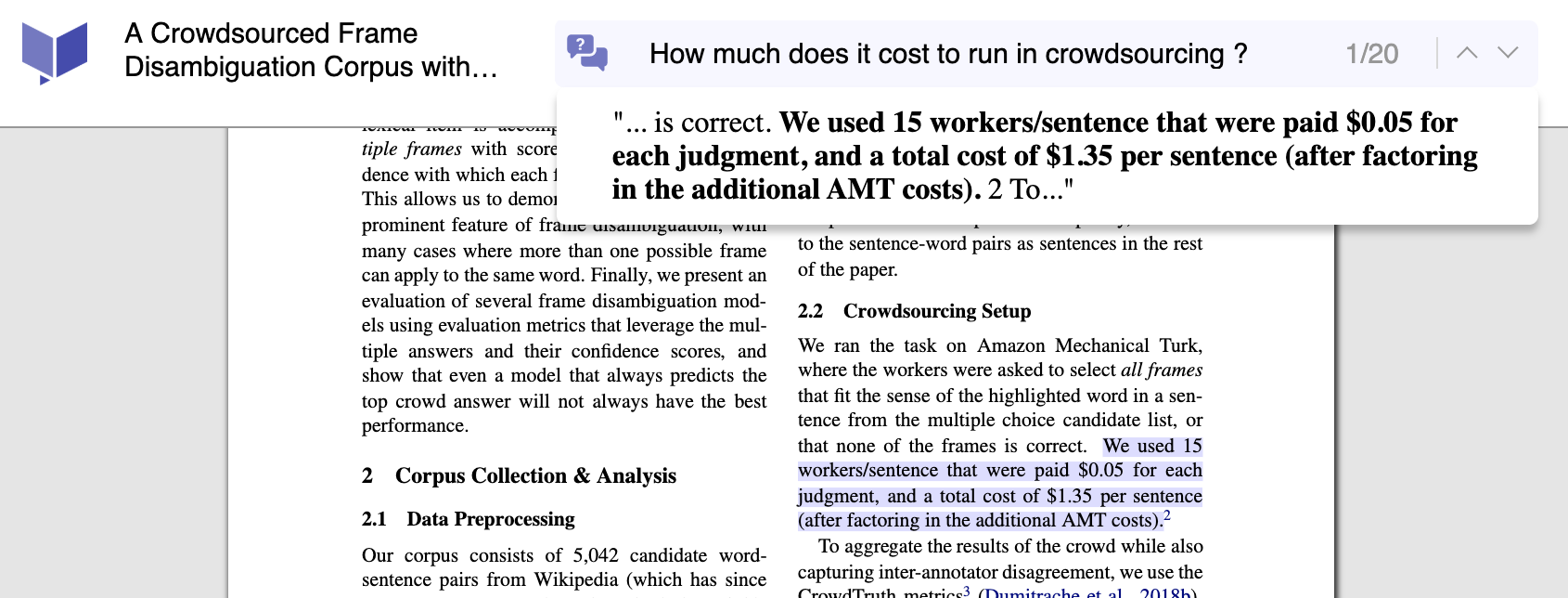}
\caption{In-paper Search Page of Talk to Paper.}
\label{fig:preview}
\end{figure*}

We first introduce SOCO (Search Oriented COnversation), which we used to build our Talk to Paper. SOCO \footnote{\url{https://docs.soco.ai/}} is an answer-engine platform that enables developers to easily build universal question answering systems with unstructured documents as its knowledge base. Figure~\ref{fig:overall} shows the overall architecture of SOCO engine. It's designed to enable users to use natural language queries to find precise answers and extract insights from massive amount of text data. The typical workflow is as following:
\begin{enumerate}
    \item Split documents into sentences and convert each sentence with its context into semantic index (i.e. a collection of answer embedding, sparse features and other semantic features).
    \item Use natural language to query the index, which first converts the query into semantic embedding and then retrieves all the high probable answers. 
    \item (Optional) Run machine reader to narrow down to phrase-level answers. 
\end{enumerate}

\subsection{SOCO-Question Answering}
We define a frame to be the basic building block of SOCO. Each frame contains $f_i=[a_i, c_i, Q_i] \quad i \in N$, where $N$ is the total number of frames, $a_i$ is the potential answer sentence, $c_i$ is surrounding context of $a_i$, and $Q_i$ is a set of questions that are manually/synthetically associated with the answer $a_i$. Note that $Q_i$ is optional and often only a small set of frames are manually labelled. 

There are two neural network models involve in SOCO QA. The first model $h_a = F_a(a, c)$ is an answer encoder that takes both the answer sentence and its surround context to create a context-sensitive answer embeddings $h_a$. The second model is a question encoder $h_q = F_q(q)$ that takes a query as input and maps it to a question embedding vector of the same size. Last, we define the relevance between a query and an answer frame to be $s = \text{cos}(h_a, h_q)$. 

\subsubsection{Training}
These two models are trained jointly via supervised learning on existing QA dataset with cross entropy loss, i.e.
\begin{equation}
    L = - \sum_{j \in J_{\text{pos}}} \log(s_j) - \sum_{j \in J_{\text{neg}}} \log(1-s_j)
\end{equation}
where $J_\text{pos}$ is the set of ground truth question-answer pairs, and $J_\text{neg}$ is the set of negative examples with randomly sampled noisy answers.

Given these two models and a set of frames, SOCO creates an index by encoding both the answers and annotated questions using $F_a$ and $F_q$ and save the resulting vectors $D$ for nearest neighbour retrieval. Since $F_q$ and $F_a$ are trained to map the input text into the same embedding space, question-to-answer relevance and question-to-question relevance can be computed and compared in the same scale via cosine similarity.

\subsubsection{Inference}
At inference stage, SOCO first encodes the input query $q'$ via $h_{q'} = F_q(q')$. Then each answer in the QA-index is scored by the cosine similarity between the query embedding and each answer embedding with a weighted auxiliary score from classic BM25 score~\cite{robertson2009probabilistic}.
\begin{multline}
    y_i = \text{cos}(h_i, h_{q'}) +\alpha \text{BM25}(a_i, q') \quad i \in |D|
\end{multline}

Note that an answer may have more than one vectors in the index because of the optional annotated question $Q$ set in the frame, i.e. $[h_a, \{h_q\}] \quad q \in Q$. We merge the scores for the same answers via max pooling. Eventually, SOCO outputs the top K answers based on the final score. 

\subsection{SOCO-Question Generation}
One common issue for new users to use question answering system is that they may not know what kind of questions they can ask. Question generation~\cite{du2017learning} is one of the solutions to this issue by suggesting users potential questions they may enter. Concretely, we created a question generator by fine-tuning a GPT-2 language model~\cite{radford2019language}.  We train the model by concatenating question answers pairs $[a, q]$ from QA corpus and fine tune a GPT-2 by maximizing the conditional log likelihood $\log P(q|a)$. The results questions are added to the $Q$ set of each frame and is used to provide auto completion and FAQs in the search interface.

\subsection{Implementation Details}
The SOCO python package (\textit{soco-core-python}) is publicly available and can be installed as a Python package by running \textit{pip3 install soco-core-python}. Internally, SOCO uses Elastic search (ES)~\cite{gormley2015elasticsearch} as its index backbone. ES has built-in support for vector search, BM25 as well as context filtering. The answer and question encoder are trained on publicly available QA datasets, including SQuAD~\cite{rajpurkar2016squad}, Natural Questions~\cite{kwiatkowski2019natural} and MSMARCO~\cite{nguyen2016ms}.


\section{Talk to Paper}
\begin{table*}[ht!]
\centering
    \begin{tabular}{|p{0.47\textwidth}|p{0.47\textwidth}|} \hline
    \textbf{Examples}   &  \textbf{Paragraphs}                 \\ \hline
    \textbf{Q:} what are pretraining objectives? \newline
    \textbf{A:} that pretraining will improve downstream tasks with fine-tuning on the entire available data \newline\newline
    \textbf{Title:} Pretraining Methods for Dialog Context Representation Learning &  ... The pretraining objectives are assessed under four different hypotheses: (1) \textbf{that pretraining will improve downstream tasks with fine-tuning on the entire available data}, (2) that pretraining will result in better convergence, ... \\ \hline
    
    \textbf{Q:} what is LSTM? \newline
    \textbf{A:} Long Short-Term Memory Network \newline\newline
    \textbf{Title:} Reasoning with Sarcasm by Reading In-between &  ... The filter width is 3 and number of filters f = 100. • LSTM is a vanilla \textbf{Long Short-Term Memory Network}. The size of the LSTM cell is set to d = 100. • ATT-LSTM (Attention-based LSTM) is a LSTM model with a neural attention mechanism applied to all the...
    
    \\ \hline
    \textbf{Q:} What is the best system for NLI? \newline
    \textbf{A:} Currently, one of the best performing NLI models (e.g., on the SNLI dataset) for three way classification is (Liu et al., 2019). \newline\newline
    \textbf{Title:} Identification of Tasks, Datasets, Evaluation Metrics, and Numeric Scores for Scientific Leaderboards Construction &  ... Our work differs in the information extracted and consequently in what context and hypothesis information we model. \textbf{Currently, one of the best performing NLI models (e.g., on the SNLI dataset) for three way classification is (Liu et al., 2019).} The authors apply deep neural networks and make use of BERT (Devlin et al., 2019),... \\ \hline
    \end{tabular}
\caption{Example results from real user queries}
\label{table:example}
\end{table*}

Now we are ready to describe the proposed Talk to Paper application, powered by our SOCO QA framework.

\subsection{Data Source}
Talk to Paper's data source contains NLP papers published last 3 years in ACL, NAACL, EMNLP and SiGdial in ACL Anthology\footnote{\url{https://www.aclweb.org/anthology/}}, which attributes to 3897 papers published in the proceedings of these conferences (we will continuously expand the database by adding more papers from previous years as well as new published papers). We first use SOCO's document parser to extract text data from the PDFs and converted them into the frame format defined in the previous section. Then we use soco-core-python to index the frames and query for answers via its RESTful API endpoint. The indexing process takes about 2 hours.

\subsection{User Interfaces}
Talk to Paper is an web app that can be used on any modern browser. There are three major pages:
\begin{itemize}
    \item Main search page
    \item In-paper search page
    \item Annotation page.
\end{itemize}

\begin{figure}[ht]
\centering
\includegraphics[width=0.45\textwidth]{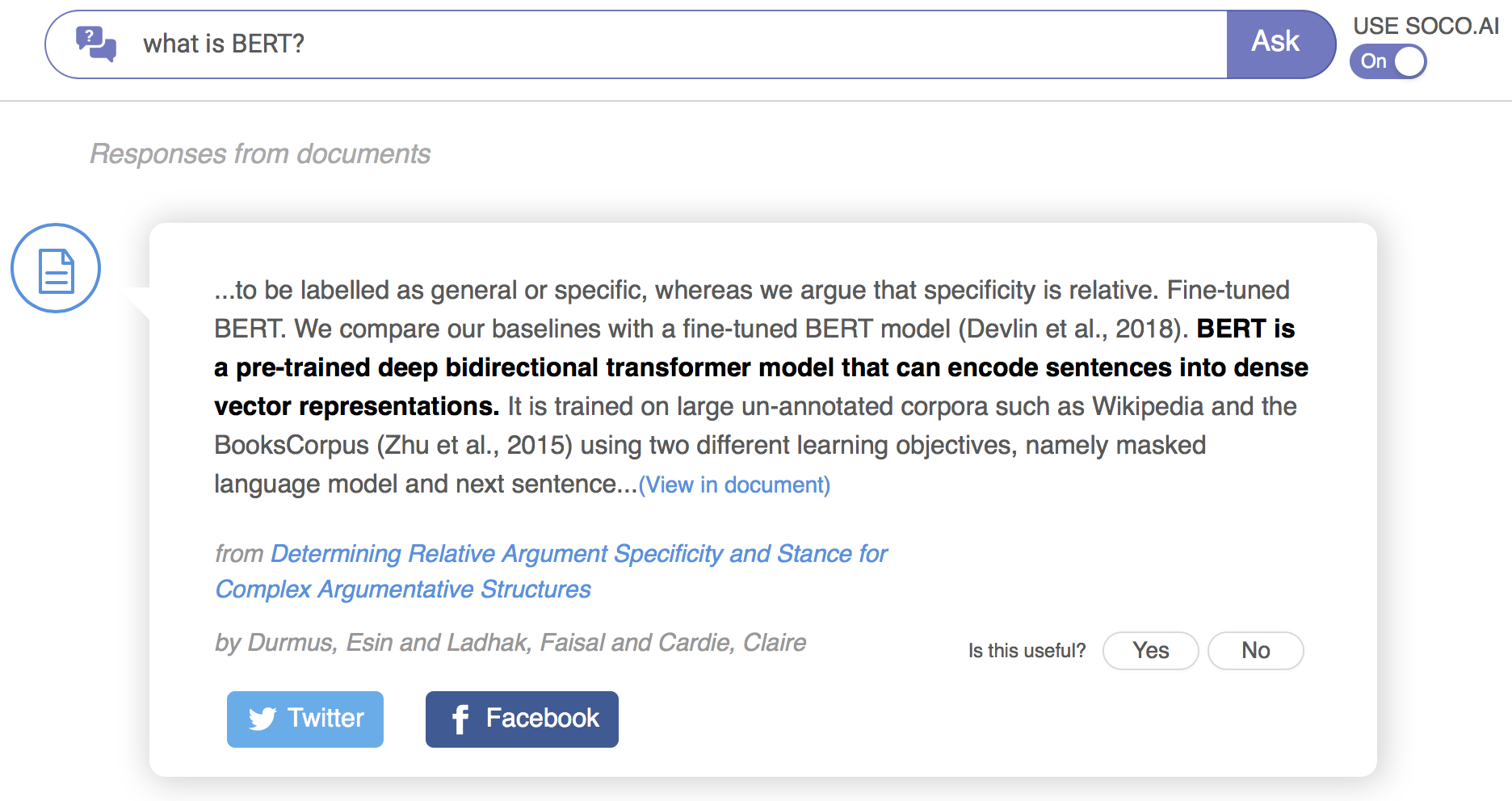}
\caption{Main search page of Talk to Paper.}
\label{fig:main}
\end{figure}

\textbf{Main Search page:} The main search page is similar to the standard Google-like search interface as shown in Figure~\ref{fig:main}, including input search box and query auto completion (based on generated questions from GPT-2).The responding answers will be highlighted in each returned results.

\textbf{In-paper Search Page:} Previously, people search information in the paper by clicking Control+F, which is a well-known shortcut key often used to find text in the current page using the exact character matching or regular expression. It is often used to input a keyword and highlight the matched string and allow to navigate the next matching or previous matching. We provide a similar interface to find the answer inside a specific paper as shown in Figure~\ref{fig:preview}. Instead of searching information using a keyword, the proposed method allow to find the information using natural language queries. The retrieved answers are highlighted and it is also allowed to navigate next answer or previous answer. It will be useful to find multiple answers in the paper. 

\begin{figure}[ht]
\centering
\includegraphics[width=0.45\textwidth]{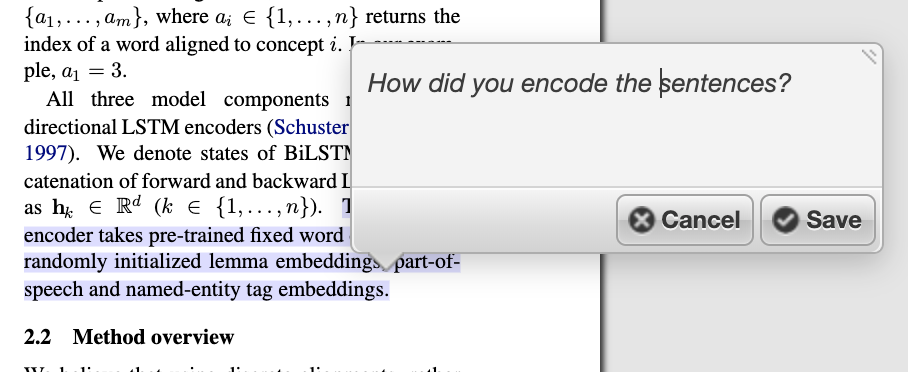}
\caption{Annotation page of Talk to Paper.}
\label{fig:annotate}
\end{figure}

\textbf{Annotation Page:} We allow to annotate the question and answer spans in the in-paper search page as shown in Figure~\ref{fig:annotate}. All annotated data are visible in the preview page. If a user wants to annotate the data, the user can simply drag the text and write a question. The data will be automatically saved in the database. Unlike other open-domain QA datasets, we cannot ask to crowd workers, students, or part-time contractors to annotate on academic papers because it is hard to annotate without the domain knowledge. Therefore, we will welcome contributions from the research community to make useful resources together for the further research. 

\subsection{Use Cases}
The typical use cases are as following: 
\begin{enumerate}
    \item A user asks a question or click one of FAQs in the main search page. The N-best results will be presented with the highlighted answer with its previous and next context. The related FAQs are also presented with the "You may also want to know" message. The user can also uses filters to narrow down to the answer in one or more specific paper.
    \item the user clicks the "view in document" to check the answer with the original paper. The in-paper search page will be shown. The user can either read the paper or uses in-paper search, e.g. \textit{what is the main contribution?} to let Talk to Paper auto scroll and highlight relevant answer spans (Figure~\ref{fig:preview}). 
    \item the user may think certain span in the paper contains important information and uses the annotation function to add related questions to this span. This new annotations will be saved in to databases and will be added to the public dataset after manual inspection.
    \item the user may uses the dataset as way to train and test performance of a question answering system. The Talk to Paper dataset is different from existing corpus because it contains highly technical text data that are substantially different from Wikipedia, which is a major source of most of the existing QA datasets.
\end{enumerate}

\section{Experiments and Results}
In this section, we first present quantitative preliminary evaluation results the effectiveness of the proposed SOCO-QA framework on a number of standard QA datasets. Then we show results on the data collected from our initial user study.

\subsection{Results for SOCO-QA performance}
This preliminary studies focuses on comparison between SOCO-QA against classic BM25~\cite{robertson2009probabilistic}. BM25-based methods remain to be the mainstream methods for document retrieval in industry. Previous work in open domain question answering has shown that BM25 is a difficult baseline to surpass when questions were written by workers who have prior knowledge of the answer~\cite{lee2019latent}. We will leave more comprehensive comparisons against other learning-based methods to future work, since the main goal of this demo paper is to present the system along with its dataset. We use the built-in elastic search~\cite{gormley2015elasticsearch} BM-25 implementation with standard English anazlyer.

\textbf{Evaluation Methods}: we compared performance on four QA datasets, i.e. SQuAD~\cite{rajpurkar2016squad}, Natural Questions (NQ)~\cite{kwiatkowski2019natural}, MS MARCO~\cite{nguyen2016ms} and Trivia QA~\cite{joshi2017triviaqa}. We break documents from the development set into sentence-level answer frames, and uses the queries in the development set to compute Mean Reciprocal Rank (MRR) and Recall at 5 (R@5) as the evaluation metrics. The data statics are summarized in Table~\ref{tbl:data}. 
\begin{table}[]
\centering
\begin{tabular}{l|ll} \hline
       & Index Size & Num of Queries \\ \hline
SQuAD  & 10,250      & 11,426          \\
NQ     & 7,020       & 1,772           \\
MARCO  & 52,933      & 13,557          \\
Trivia & 26,345      & 8,165           \\ \hline
\end{tabular}
\caption{Statistics on the evaluation datasets.}
\label{tbl:data}
\end{table}

\textbf{Quantitative Results:} 
\begin{table}[]
\centering
\begin{tabular}{l|ll|ll} \hline
       & BM25 & & SOCO \\
       & MRR  & R@5 & MRR  & R@5 \\ \hline
SQuAD  & 58.0 & 69.0 & \textbf{60.9} & \textbf{73.2}\\
Trivia & 29.0 & 38.7 & \textbf{34.0} & \textbf{59.2}\\
NQ     & 19.7 & 25.1 & \textbf{69.3} & \textbf{87.3} \\
MARCO  & 20.7 & 27.0 & \textbf{73.2} & \textbf{92.8} \\ \hline
\end{tabular}
\caption{Main evaluation results.}
\label{tbl:mrr}
\end{table}
Table~\ref{tbl:mrr} shows the main results. The proposed SOCO-QA model is able to significantly outperform the baseline BM25 on all datasets. The proposed method is particularly powerful on real query data, e.g. NQ and MARCO where the question writer does not the exact answer they are looking for, so that there is often a low word overlapping between the question and the answer. Table~\ref{tbl:mrr} shows a striking $251\%$ and $253.6\%$ relative MRR improvement on the NQ and MARCO dataset. On the other hand, SOCO is also able to beat BM25 on SQuAD and Trivia dataset, where there is significant more question-to-answer word lapping.

\textbf{Qualitative Results:} to provide better understanding between BM25-based search versus SOCO-QA, the following are some example side-by-side comparisons:
\begin{itemize}
    \item \textbf{SOCO:} We compare our baselines with a fine-tuned BERT model (Devlin et al., 2018). \textbf{BERT is a pre-trained deep bidirectional transformer model that can encode sentences into dense vector representations.} It is trained on large un-annotated corpora such as Wikipedia and the BooksCorpus (Zhu et al., 2015).
    \item \textbf{ES Default (BM25):} for the claim pairs with distance values 2 to 5 as shown in Table 3. We find \textbf{that BERT} model \textbf{is} consistently the \textbf{best} performing model for all distance pairs. As we increase the distance, the models achieve higher prediction performance.
\end{itemize}
The main observations is that BM25 falls short in understanding the intent of the query. Although it is also able to find sentences that are relevant to the query terms, it does not rank sentences that can ``answer'' the query higher. On the other hand, SOCO-QA is able to recognize target answer a query is looking for, e.g. a definition, and rank sentences that can directly resolves the questions higher. 

\subsection{Data Analysis}
We asked NLP researchers via social network, e.g. Twitter, to try out Talk to Paper and we are able to collect 3137 queries in roughly two weeks. The logged query data and its annotation will soon be made publicly available). Table~\ref{table:example} shows example queries where the system is able to find relevant answers to real user queries. Analysis shows that the most frequent query type were asking about the objectives or the meaning of terms (e.g., what are pretraining objectives, what is LSTM?). Another popular question type is to ask about the state-of-the-art method to solve a particular problem, e.g. What is the best system for NLI?. 

We also found that the generated questions that are presented as auto-completion and FAQs are particularly popular. About 51.7\% of queries were from the suggested questions. This results is inline with research work in human-computer interaction that utilizes machine intelligent systems to assist human users to better discover knowledge~\cite{lee2019learning}. 

\section{Conclusion}
We present Talk to Paper, a QA system for NLP papers powered by SOCO-QA. Experiments confirm the effectiveness of our proposed approach and show superior search experience compared to traditional search engine. We welcome contributions from the research community to curate useful resources together for the further research. Future work include (1) expanding the database to more papers (2) improving the QA model using the collected data to better handle question answering in the context of research domain.  
\newpage

\section*{Acknowledgments}
We would like to acknowledge the joint effort from SOCO's development team, including Haolin Wang, Yanran Han and  Omer Riaz to make this work possible.

\bibliography{acl2020}
\bibliographystyle{acl_natbib}

\appendix

\end{document}